\documentclass[11pt]{article}
\usepackage[margin=1in]{geometry}
\usepackage{amsmath,amssymb}
\usepackage{booktabs}
\usepackage{algorithm}
\usepackage{algpseudocode}
\usepackage{graphicx}
\usepackage{tikz}
\usetikzlibrary{arrows.meta,positioning,fit,calc}
\usepackage{hyperref}
\usepackage[numbers,sort&compress]{natbib}

\title{Structured Affinity for Unsupervised Visual Class-Incremental Memory in Deep Artificial Immune Networks}
\author{Siphesihle Sithungu\\[0.75em]
University of Johannesburg\\
Johannesburg, South Africa\\
siphesihles@uj.ac.za}
\date{}

\begin{document}

\maketitle

\begin{abstract}
Artificial immune networks (AINs) are naturally memory-forming systems, but conventional visual AINs often rely on flattened vector affinity that ignores spatial structure. This paper studies whether structured, gradient-free immune affinity can make Deep AINs viable as replay-free visual class-incremental representation-memory learners. Visual B-cells are formalized as structured templates, including shifted-template affinity, zero-normalized cross-correlation (ZNCC) filters, and feature-map binding profiles. A repertoire is treated both as memory and as a representation-inducing basis, while depth is obtained by passing binding-profile response maps to subsequent immune layers. The resulting Deep AIN exhibits adaptive latent coordinate reorganization: as new classes arrive, the binding-profile space evolves while retaining recoverable structure for earlier classes. Experiments on sklearn digits, MNIST, Fashion-MNIST, and KMNIST show that preserving response maps is critical. Scalar binding-profile variants underperform, whereas feature-map Deep AINs learn class-discriminative visual memory without replay, label-driven immune updates, or backpropagation through the immune layers. On sklearn digits, downstream probes fitted on the learned binding profiles reach 0.939 final balanced accuracy with logistic regression and 0.902 with 1-nearest-neighbour after all ten classes are encountered, with initial-class retention of 0.978. Adaptive layer-wise scale calibration further improves the two-layer feature-map Deep AIN to 0.978 balanced accuracy. With the same calibration rule, Fashion-MNIST reaches 0.814 and KMNIST reaches 0.853. These probes are external validation tools, not components of the AIN. The results identify structured affinity, response-map preservation, adaptive latent reorganization, and layer-wise scale calibration as key mechanisms for replay-free visual immune memory.
\end{abstract}


\section{Introduction}

Artificial immune networks are naturally suited to memory-centred learning: they encounter antigens, adapt repertoires, retain useful responses, and discriminate among previously encountered patterns. However, visual data exposes a limitation of many conventional AIN formulations. If an image is treated as a flattened vector, the affinity between an antigen and a B-cell ignores spatial locality, translation, and local visual structure. This can make the AIN appear weak even when the underlying difficulty is the affinity operator rather than immune memory itself.

This paper studies visual AINs from the standpoint of online memory formation. A class arrives, the repertoire forms memory for that class, later classes arrive, and the system must retain and discriminate among all encountered classes. At stream step $t$, the model has encountered classes $\mathcal{C}_t$, and evaluation is performed only over $\mathcal{C}_t$. This setting directly tests whether an immune repertoire can acquire new visual experience while preserving earlier memory.

The central hypothesis is that AINs become viable visual online learners when affinity is made compatible with image structure. Instead of representing a B-cell only as a point in a flattened pixel space, a visual B-cell can be a structured template or local filter. Its response to an image can form part of a binding profile, and stacked binding profiles can define a Deep AIN in which later layers learn over earlier immune responses. This makes the repertoire both a memory population and a representation-inducing basis.

The paper makes five contributions:

\begin{enumerate}
    \item \textbf{Structured visual immune affinity.} The paper defines image-aware B-cell affinity operators that remain gradient-free, including shifted-template affinity, ZNCC filters, and feature-map response profiles.
    \item \textbf{A visual class-incremental memory protocol.} The paper evaluates AINs on sequential class acquisition, old-class retention, balanced accuracy over encountered classes, memory growth, and update cost.
    \item \textbf{Deep visual binding profiles.} The paper defines depth as the composition of visual response maps and binding-profile maps, so later immune layers learn memories over earlier immune responses.
    \item \textbf{Empirical evidence for online binding-profile memory.} The paper shows that an online feature-map Deep AIN can learn binding-profile memory spaces whose class structure is recoverable by simple downstream probes, without using labels to update the immune layers.
    \item \textbf{A stability-plasticity diagnosis for visual AINs.} The paper separates memory-slot growth from lower-representation adaptation, showing when adapting lower visual filters helps and when it destabilizes later memory.
\end{enumerate}

\section{Similar Works}

\subsection{Artificial Immune Systems and Immune Memory}

Artificial immune systems are computational models inspired by immune recognition, clonal expansion, mutation, suppression, memory formation, and self/non-self discrimination \citep{farmer1986immune,dasgupta1999ais,decastro2002ais}. Immune network theory is particularly relevant to this paper because it views immune behaviour as an adaptive population process rather than as the optimization of a single global parameter vector \citep{jerne1974network}. Artificial immune networks inherit this view by maintaining a repertoire of antibodies or B-cells whose affinities determine stimulation, adaptation, survival, and suppression.

Several AIS families are related to the present work. Negative selection algorithms construct detectors for self/non-self discrimination \citep{forrest1994negative}. Clonal selection algorithms adapt high-affinity candidates through cloning and mutation \citep{decastro2002clonal}. Artificial immune recognition systems such as AIRS use immune-inspired memory cells for supervised classification \citep{watkins2002airs}. aiNet-style artificial immune networks use immune interaction and suppression for clustering, compression, and data analysis \citep{decastro2000ainet,timmis2000resource}. These methods establish immune memory as a computational learning mechanism, but they usually operate over vector features and are not primarily formulated as layered visual representation learners. The present paper keeps the immune-memory interpretation, but asks how affinity must be structured when (1) the antigen is an image, (2) learning occurs in a class-incremental stream, and (3) the network should not only learn the feature space but must also learn a latent space over binding profiles.

\subsection{Prototype, Kernel, and Dissimilarity Representations}

The binding-profile representation used here is closely related to prototype and similarity-based learning. RBF networks represent an input by its response to a set of centres \citep{broomhead1988rbf,bishop1995nn}. Learning vector quantization and self-organizing maps use learned reference vectors to organize data and support classification \citep{kohonen1982som,kohonen1990lvq}. Dissimilarity representations describe an object by its dissimilarities to a representation set rather than by its original coordinates \citep{pekalska2005dissimilarity}. Nystr\"om methods and random Fourier features similarly construct useful feature spaces from landmark or kernel-induced responses \citep{williams2001nystrom,rahimi2007random}.

This paper does not claim that representing an object by responses to reference patterns is new. The contribution is the immune realization of this idea for online visual memory. The reference set is an adaptive B-cell repertoire, binding profiles are the response coordinates, and the representation is updated by immune operations rather than by gradient descent. This distinction matters in the class-incremental setting because memory formation and representation formation are coupled: adding or adapting B-cells changes both what is remembered and the coordinates in which future antigens are represented.

\subsection{Structured Visual Matching}

Flattening an image into a vector discards spatial locality and makes small translations appear as large coordinate changes. Classical computer vision addressed this problem using structured local matching, local descriptors, and translation-tolerant template comparison. Normalized cross-correlation and its efficient variants are standard tools for template matching because they compare local patterns after compensating for mean and contrast differences \citep{lewis1995fast}. Local descriptor methods such as Scale-Invariant Feature Transform (SIFT) and Histogram of Oriented Gradients (HOG) further show that robust visual recognition often depends on local spatial structure rather than raw pixel-vector distance \citep{lowe2004distinctive,dalal2005hog}.

Convolutional neural networks also exploit locality and weight sharing, and have become the dominant learned approach for visual recognition \citep{fukushima1980neocognitron,lecun1998gradient}. The structured affinity method proposed here is related to convolution in the operational sense that a local B-cell filter is evaluated across image patches. However, it is not a neural convolutional layer: filters are B-cells, responses are immunonological affinities, and adaptation occurs through immune repertoire dynamics rather than backpropagation. The purpose is therefore not to replace convolutional networks as supervised visual classifiers, but to test whether immune repertoires can form useful online visual memory when their affinity operator respects image structure.

\subsection{Online, Incremental, and Continual Visual Learning}

The experimental setting is also related to online and continual learning. Continual learning studies how a learner can acquire new information without catastrophically forgetting previous knowledge \citep{parisi2019continual,delange2021continual,vandeven2022continual}. Neural continual-learning methods include regularization approaches such as elastic weight consolidation \citep{kirkpatrick2017ewc}, exemplar or replay-based methods such as Incremental Classifier on Representation Learning (iCaRL) \citep{rebuffi2017icarl}, and constrained update methods such as gradient episodic memory \citep{lopezpaz2017gem}. These approaches usually target supervised predictive performance and often rely on stored examples, replay buffers, or gradient-based parameter updates.

The present paper is positioned differently. It evaluates whether an AIN can learn an online visual representation-memory space without replay and without backpropagation through the immune layers. Downstream classifiers are used only as probes of the learned binding-profile space, analogous to representation evaluation protocols. The immune model itself is not trained to minimize classification loss. This makes the comparison to replay MLPs and refit baselines informative but not defining: those baselines indicate how much supervised rehearsal or refitting can buy, whereas the AIN result tests whether structured immune memory can produce a useful visual representation under sequential exposure.

\subsection{Positioning of This Paper}

The closest conceptual neighbourhood of this work is the intersection of AIN memory, prototype-response representations, structured visual matching, and continual learning. The paper's specific claim is narrower than general visual classification and broader than a single affinity ablation. It argues that visual AINs become viable online memory learners when the affinity mechanism is made compatible with the structure of the antigen. In this view, a B-cell is not merely a point in a feature space; it can be a local visual template whose response contributes to a spatial binding profile. A Deep AIN then learns over those response profiles, allowing later immune layers to model patterns of earlier immune responses.

This positioning separates the proposed model from four adjacent lines of work. Unlike conventional AIS classifiers, the immune layers are treated as representation-memory learners rather than as a direct supervised classifier. Unlike ordinary prototype or kernel methods, the representation basis is formed by immune adaptation and can grow or change online. Unlike convolutional neural networks, the visual filters are not optimized by backpropagation. Unlike standard continual-learning baselines, the central object of study is the learned immune binding-profile space rather than a replay-trained discriminative model.

\section{Model}

The model is built around a simple immune-representation idea: a B-cell repertoire is both a memory population and a coordinate system. When an antigen is presented to a repertoire, it does not only stimulate individual B-cells; it also produces a pattern of responses across the full repertoire. That response pattern is the antigen's binding profile. A Deep AIN is obtained by feeding one layer's binding-profile response into the next layer as its antigenic input.

\begin{figure}[htbp]
\centering
\begin{center}
\includegraphics[width=0.92\textwidth,height=0.32\textheight,keepaspectratio]{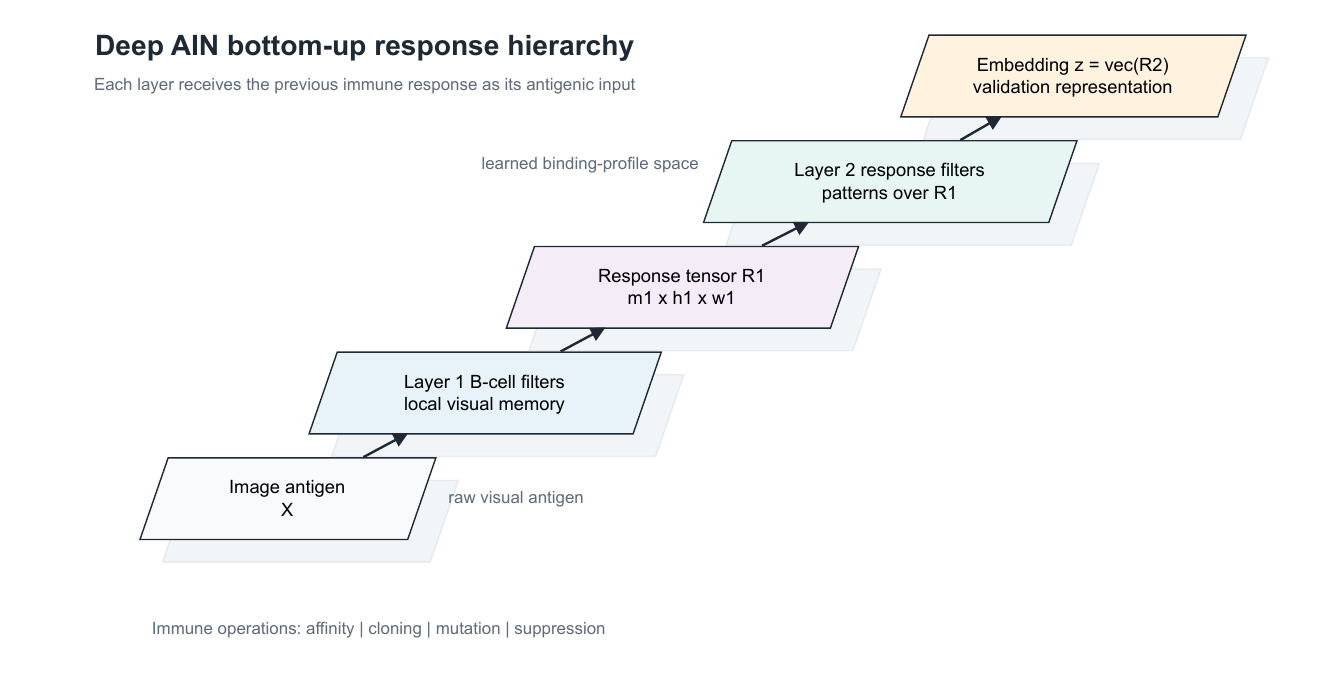}
\end{center}
\caption{High-level Deep AIN construction. Each layer transforms its antigenic input into a repertoire response. Later layers learn over earlier response patterns.}
\label{fig:high_level_deep_ain}
\end{figure}

Figure~\ref{fig:high_level_deep_ain} shows the high-level construction. The first layer receives an image. Its B-cells respond to local visual structure and produce response maps. The second layer receives the stack of response maps and learns response-pattern memory. Further layers, if used, repeat the same principle: each layer receives the previous layer's immune response, not an externally engineered feature vector.

\subsection{B-Cell Repertoires and Binding Profiles}

Let a repertoire at layer $\ell$ be
\[
    \mathcal{B}^{(\ell)} = \{b^{(\ell)}_1,\ldots,b^{(\ell)}_{m_\ell}\},
\]
where each $b^{(\ell)}_i$ is a B-cell template in that layer's antigen space. For an input $u^{(\ell-1)}$, the layer computes affinities between the input and every B-cell. In the scalar case, the binding profile is
\[
    \Phi_{\mathcal{B}^{(\ell)}}(u^{(\ell-1)})
    =
    \left[
    K(u^{(\ell-1)}, b^{(\ell)}_1),
    \ldots,
    K(u^{(\ell-1)}, b^{(\ell)}_{m_\ell})
    \right].
\]
This vector is a coordinate representation induced by the repertoire. A B-cell therefore has two roles: it is a memory element, and it defines one coordinate of the binding-profile representation.

For non-visual data, this construction can be implemented with a standard RBF affinity,
\[
    K_{\mathrm{flat}}(x,b) =
    \exp\left(-\frac{\|x-b\|^2}{2\sigma^2}\right).
\]
For images, however, this flattened affinity is a weak inductive bias because it treats an image as a coordinate vector rather than as a spatial object. The rest of the model replaces this flat affinity with structured visual affinity.

\subsection{Structured Visual Affinity}

The first structured variant keeps B-cells as full-image templates but allows small translations before computing affinity. For image $X$ and template $B$,
\[
    K_{\mathrm{shift}}(X,B) =
    \max_{\Delta \in \mathcal{T}_r} K_{\mathrm{flat}}(X, T_{\Delta}B),
\]
where $\mathcal{T}_r$ is a local translation window and $T_{\Delta}$ applies a zero-padded shift. This gives the B-cell limited translation tolerance while keeping the immune update mechanism unchanged.

The main visual model uses convolution-style ZNCC affinity. A B-cell can be a local $k\times k$ visual filter. For an image patch $P$ and B-cell filter $b$, the local normalized correlation is
\[
    \mathrm{ZNCC}(P,b) =
    \frac{\langle P-\bar{P}, b-\bar{b}\rangle}
    {\|P-\bar{P}\|\|b-\bar{b}\|}.
\]
The response is mapped to $[0,1]$:
\[
    K(P,b) = \frac{\mathrm{ZNCC}(P,b)+1}{2}.
\]
Sliding the B-cell over all valid image patches produces a response map. If the response is collapsed to a scalar, the affinity of filter $b$ to image $X$ is
\[
    K_{\mathrm{filter}}(X,b) = \max_{P \in \mathcal{P}_k(X)} K(P,b).
\]
This operation is convolution-style in the computer-vision sense, but it is not a neural convolutional layer. Filters are B-cells, responses are immune affinities, and the filters are adapted by immune operations rather than backpropagation.

\begin{figure}[htbp]
\centering
\begin{center}
\includegraphics[width=0.92\textwidth,height=0.32\textheight,keepaspectratio]{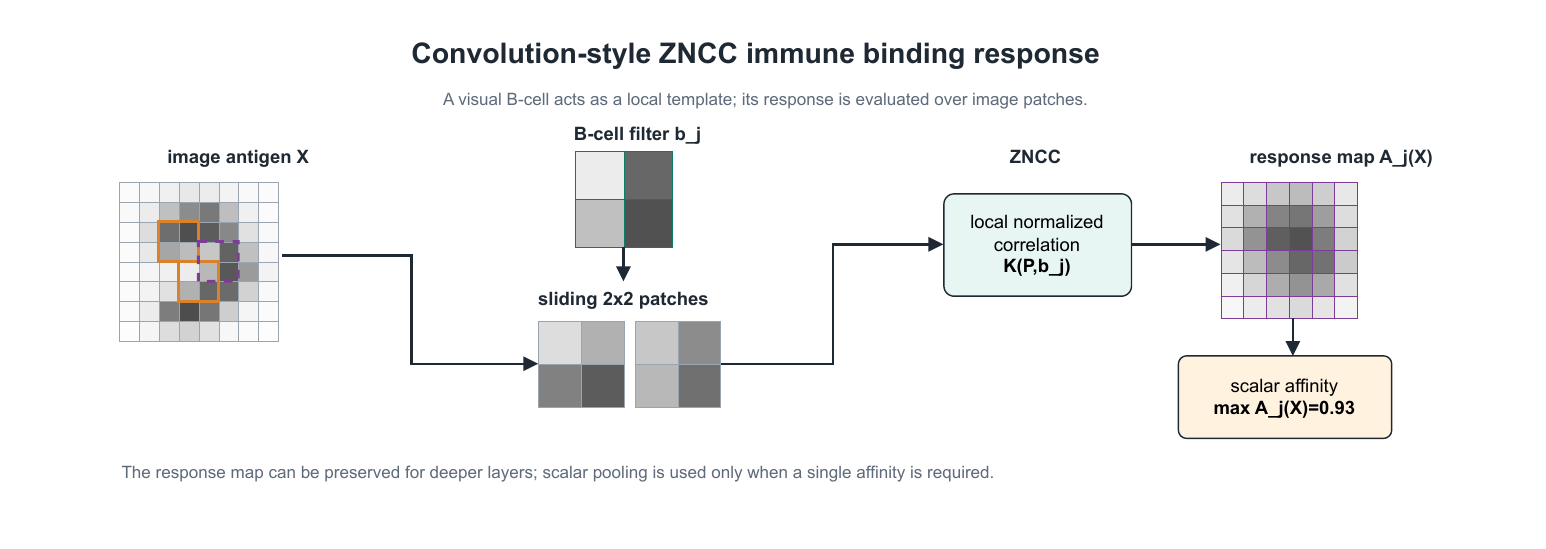}
\end{center}
\caption{Convolution-style ZNCC binding response. A local B-cell filter is evaluated over image patches, producing a response map. The response map may be preserved for deeper layers or pooled into a scalar affinity.}
\label{fig:conv_zncc_profile}
\end{figure}

\subsection{Feature-Map Deep AIN}

The feature-map Deep AIN preserves the spatial response maps instead of immediately collapsing each B-cell response to a scalar. If layer 1 has $m_1$ local visual B-cells, it maps an image to
\[
    R^{(1)}(X) \in \mathbb{R}^{m_1 \times h_1 \times w_1}.
\]
This tensor is a stack of immune response maps, one map per layer-1 B-cell. Layer 2 receives this response-map stack as its antigenic input. A layer-2 B-cell is therefore not an image patch; it is a filter over lower-layer response maps. With $m_2$ layer-2 B-cells, the second response is
\[
    R^{(2)}(X) \in \mathbb{R}^{m_2 \times h_2 \times w_2}.
\]
The validation embedding used in the experiments is
\[
    z(X)=\mathrm{vec}(R^{(2)}(X)).
\]

More generally, a Deep AIN composes immune response maps:
\[
    R^{(\ell)}(X) =
    \Phi_{\mathcal{B}^{(\ell)}}(R^{(\ell-1)}(X)),
    \qquad R^{(0)}(X)=X.
\]
The important point is that depth is not obtained by attaching a classifier to an AIN. Depth is obtained by making the previous immune response the next immune layer's antigenic input.

\subsection{Online Immune Updating}

Each repertoire is updated online using immune operations: affinity evaluation, stimulation of responsive B-cells, clonal expansion, mutation, and suppression. The immune update does not use class labels for the feature-map layers. Labels are used only by the external downstream validation probes after embeddings have been produced. This separation is essential: the AIN learns a visual memory representation, and the probe measures whether that representation contains recoverable class structure.

When a new class batch arrives, the lower visual repertoire may adapt, and the binding-profile coordinate system may change. This adaptive coordinate reorganization is a core property of the model: the current repertoire defines the current coordinate system. Earlier samples can be re-embedded through the updated repertoire to determine whether the evolved immune memory still represents previous classes. Successful retention therefore does not require stale coordinates to remain fixed; it requires earlier classes to remain recoverable in the current binding-profile space.

\subsection{End-to-End Pipeline}

\begin{figure}[htbp]
\centering
\begin{center}
\includegraphics[width=0.92\textwidth,height=0.32\textheight,keepaspectratio]{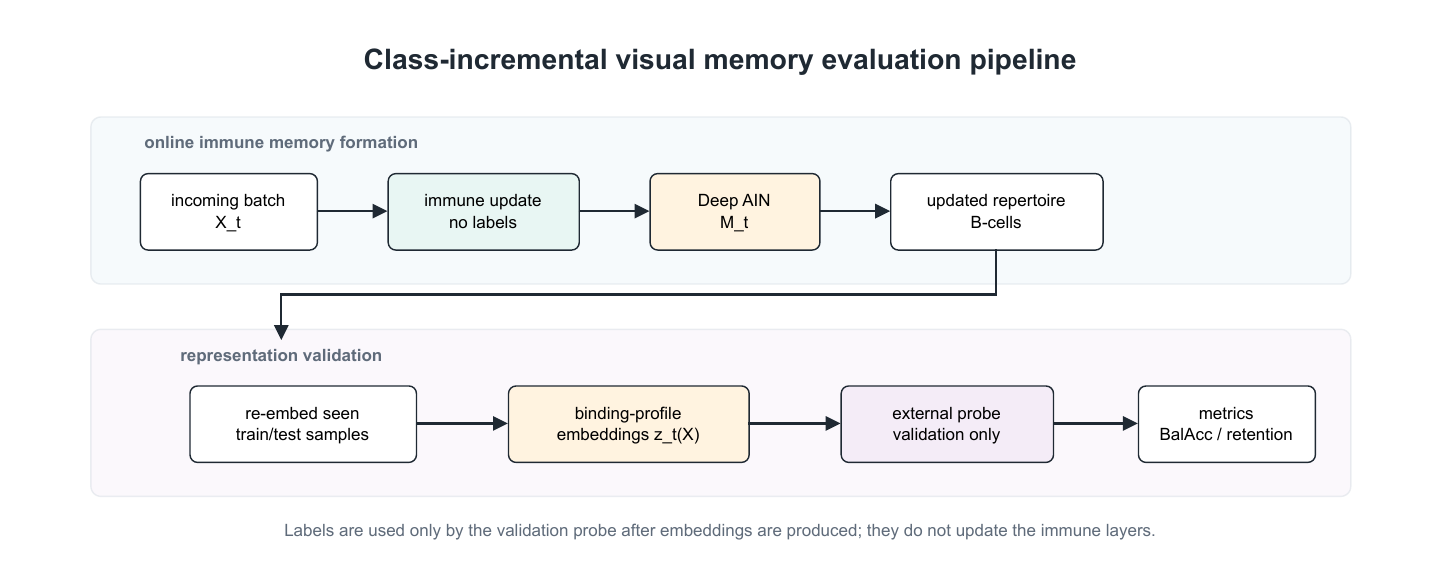}
\end{center}
\caption{Experimental pipeline. The immune layers are updated online from incoming batches. External probes are fitted only after embeddings are produced, to evaluate the current binding-profile space.}
\label{fig:model_pipeline}
\end{figure}

Figure~\ref{fig:model_pipeline} summarizes the full pipeline used in the experiments. Incoming visual batches update the immune repertoires. At evaluation time, encountered training and test samples are embedded through the current Deep AIN. External probes are then fitted on the current training embeddings and evaluated on current test embeddings. The probes validate the learned immune representation; they are not part of the immune update.

\section{Experiment Setup}

\subsection{Datasets}

The experiments use four grayscale visual benchmarks: sklearn digits, MNIST, Fashion-MNIST, and KMNIST. Sklearn digits provides $8\times8$ handwritten digit images and is used for the most detailed method comparison, timing analysis, trajectory analysis, and depth calibration. MNIST, Fashion-MNIST, and KMNIST provide $28\times28$ images and are used to test whether the same feature-map Deep AIN protocol transfers beyond the small sklearn digits setting. MNIST evaluates handwritten digit memory at a larger scale; Fashion-MNIST changes the visual domain to clothing categories; KMNIST introduces more complex character shapes and is used to diagnose the geometry of the learned binding-profile space.

All datasets are treated as class-incremental streams. The first class is used to initialize memory, and the remaining classes are introduced one at a time. The model is evaluated only on classes that have already been encountered. This focuses the experiment on the intended question: whether an immune repertoire can acquire, retain, and organize encountered visual classes as the stream grows.

\subsection{Class-Incremental Protocol}

Let $\mathcal{C}_t$ denote the set of classes encountered by stream step $t$. The protocol is:

\begin{algorithm}[htbp]
\caption{Class-incremental visual memory protocol}
\label{alg:class_incremental_protocol}
\begin{algorithmic}[1]
\Require Initial class set $\mathcal{C}_0$, stream classes $(c_1,\ldots,c_T)$, initial samples $X_0$, stream batches $(X_1,\ldots,X_T)$, test set $X^{\mathrm{test}}$
\State Initialize immune memory $M_0$ using $X_0$
\State Set encountered classes $\mathcal{C}_{\mathrm{seen}} \gets \mathcal{C}_0$
\State Evaluate $M_0$ on test samples with labels in $\mathcal{C}_{\mathrm{seen}}$
\For{$t = 1$ to $T$}
    \State Receive new class batch $X_t$ from class $c_t$
    \State Update immune memory: $M_t \gets \mathrm{ImmuneUpdate}(M_{t-1}, X_t)$
    \State Update encountered classes: $\mathcal{C}_{\mathrm{seen}} \gets \mathcal{C}_{\mathrm{seen}} \cup \{c_t\}$
    \State Embed encountered training samples through current memory $M_t$
    \State Embed encountered test samples through current memory $M_t$
    \State Fit external validation probe on current training embeddings
    \State Evaluate over test samples with labels in $\mathcal{C}_{\mathrm{seen}}$
\EndFor
\end{algorithmic}
\end{algorithm}

Unless otherwise stated, results are averaged over three random seeds. The sklearn digits, MNIST, Fashion-MNIST, and KMNIST class-incremental runs use the same stream structure: digit or class 0 initializes memory, and classes 1 through 9 arrive sequentially. The reported final-step results therefore evaluate performance after all ten classes have been encountered. The standard stream uses 80 initial samples from the first class, 60 samples per new class event, and 35 held-out test samples per class.

\subsection{Feature-Map Deep AIN Configuration}

The main proposed model is the feature-map Deep AIN. Layer 1 contains local visual B-cell filters. For a grayscale image $X$, each layer-1 B-cell produces a ZNCC response map by sliding over local image patches. Layer 2 receives the stack of layer-1 response maps as its antigenic input and learns response-map filters over that stack. The final representation used for validation is
\[
    z(X)=\mathrm{vec}(R^{(2)}(X)).
\]

The default two-layer feature-map model uses 30 layer-1 filters and 30 layer-2 filters. For $28\times28$ datasets, the first-layer filter size is $5\times5$ and the second-layer response-map filter size is $2\times2$, producing a final binding-profile embedding of dimension 15870. For sklearn digits, the same conceptual architecture is applied to $8\times8$ images, producing a smaller final representation. Immune updates use clonal selection style repertoire training with mutation and suppression; class labels are not used to update the immune layers.

Higher-layer affinity scales are important because response-map and binding-profile spaces do not have the same distance scale as the original image space. The main cross-dataset feature-map experiments therefore use an adaptive scale rule based on nearest-neighbour distances in the relevant response-profile space. The scale estimate is multiplied by a fixed factor and kept consistent across datasets in the main cross-dataset comparison. Manual fixed scales are used in the depth-calibration experiment to demonstrate the effect of scale mismatch.

\subsection{Validation Probes}

The feature-map Deep AIN is evaluated as an unsupervised online immune representation-memory learner. After each stream event, encountered training samples and encountered test samples are passed through the current immune layers to obtain binding-profile embeddings. External probes are then fitted on the encountered training embeddings and evaluated on the encountered test embeddings.

The primary probes are logistic regression and 1-nearest-neighbour. Logistic regression tests whether the learned immune representation is approximately linearly separable. The 1-nearest-neighbour probe tests local class coherence in the binding-profile space. For KMNIST, additional probes are used to diagnose geometry: nearest-centroid probing tests compact global class means, and PCA+RBF-SVM tests whether a simple nonlinear boundary improves after dimensionality reduction. These probes are validation tools, not components of the AIN. They do not update the immune repertoire and should be interpreted as measurements of the representation produced by the AIN.

This validation protocol also clarifies how adaptive coordinate reorganization is handled. The immune coordinate system may change as new classes arrive. The experiments therefore do not reuse stale embeddings. Instead, all encountered samples are re-embedded through the current AIN at each evaluation step. This tests whether the current immune memory space still represents earlier classes after online adaptation.

\subsection{Baselines}

The baseline set separates non-updating, online-updating, refit, supervised-prototype, and neural-replay behaviours:
\begin{itemize}
    \item \textbf{Static k-means}: k-means fitted to the initial class only and never updated.
    \item \textbf{Mini-batch k-means}: k-means centroids updated online with incoming class batches.
    \item \textbf{Full refit k-means}: k-means refitted on all accumulated encountered data after each class event.
    \item \textbf{Online prototype memory}: a supervised nearest-prototype memory updated toward samples from the corresponding class.
    \item \textbf{No-replay MLP}: a neural classifier updated on the current batch only.
    \item \textbf{Replay MLP}: a neural classifier trained with a replay buffer of previous examples.
\end{itemize}

The comparison also includes scalar binding-profile AIN variants in the static diagnostic: flat RBF AIN, shifted-template RBF AIN, convolution-style ZNCC AIN, and multi-ZNCC AIN. These variants test whether changing the affinity operator in the first place improves the AIN response space before moving to feature-map response preservation.

\subsection{Evaluation Metrics}

The primary metric is balanced accuracy over encountered classes. This is preferred over ordinary accuracy because class-incremental streams can produce uneven difficulty across old and new classes. Initial retention measures accuracy on the initial class after later classes have arrived. Current-class accuracy measures acquisition of the most recently introduced class. These two metrics separate stability from plasticity: a method can learn the new class while forgetting the first class, or retain the first class while failing to acquire the new class.

The experiments also report representation dimension or memory size. For prototype and k-means methods, this is the number of memory slots or centroids. For feature-map Deep AIN probes, it is the dimensionality of the final binding-profile embedding $z(X)$ used by the external probe. Runtime is reported for the sklearn digits protocol as an implementation-level diagnostic, separating immune memory training, probe fitting, total training time, and inference time.

\section{Results}

\subsection{Static Representation Diagnostic}

Static multiclass classification is not the main task, but it verifies whether image-aware affinity improves the health of the AIN response space. Table~\ref{tab:multiclass_affinity} shows that structured affinity (all the methods shown in Table~\ref{tab:multiclass_affinity} except the Flat RBF AIN) substantially improves digit recognition on sklearn digits. The MNIST feature-map result is lower but confirms that the representation is not limited to the small $8\times8$ setting.

\begin{table}[htbp]
\centering
\caption{Multiclass digit diagnostics. Sklearn digits rows report accuracy and balanced accuracy. The MNIST feature-map row reports multiclass accuracy over a 350-example-per-class subset.}
\label{tab:multiclass_affinity}
\begin{tabular}{llcc}
\toprule
Dataset & Method & Accuracy & Balanced accuracy \\
\midrule
Sklearn digits & Flat RBF AIN & 0.849 & 0.848 \\
Sklearn digits & Shifted-template RBF AIN & \textbf{0.917} & \textbf{0.917} \\
Sklearn digits & Convolution-style ZNCC AIN & \textbf{0.931} & \textbf{0.931} \\
Sklearn digits & Feature-map Deep AIN & \textbf{0.850} & -- \\
MNIST & Feature-map Deep AIN & \textbf{0.731} & -- \\
\bottomrule
\end{tabular}
\end{table}

\subsection{Class-Incremental Visual Memory}

Table~\ref{tab:class_incremental_final} reports the final step of a sklearn digits class-incremental stream. The model starts from digit 0 and then receives digits 1 through 9 sequentially. Evaluation at the final step is over all ten encountered classes. The last two rows (Feature-map Deep AIN + logistic probe and Feature-map Deep AIN + 1NN probe) are downstream probes on the learned Deep AIN binding-profile space.

\begin{table}[htbp]
\centering
\caption{Final ten-class performance in the sklearn digits class-incremental visual-memory stream. Values are averaged over three seeds. Probe rows externally validate the learned Deep AIN binding-profile memory.}
\label{tab:class_incremental_final}
\resizebox{\textwidth}{!}{%
\begin{tabular}{llcccc}
\toprule
Method & Evaluation mode & Balanced accuracy & Initial retention & Current-class accuracy & Memory size / embedding dimension \\
\midrule
Static k-means & no update & 0.100 & 1.000 & 0.000 & 80 \\
Mini-batch k-means & online centroid update & 0.617 & 0.978 & 0.678 & 80 \\
Full refit k-means & refit on seen data & 0.928 & 1.000 & 0.867 & 120 \\
Online prototype memory & supervised online prototypes & 0.872 & 1.000 & 0.800 & 120 \\
No-replay MLP & current batch only & 0.827 & 0.922 & 0.767 & -- \\
Replay MLP & replay classifier & 0.953 & 0.989 & 0.933 & -- \\
Feature-map Deep AIN + logistic probe & external probe on $z(X)$ & \textbf{0.939} & \textbf{0.978} & \textbf{0.933} & \textbf{270} \\
Feature-map Deep AIN + 1NN probe & external probe on $z(X)$ & \textbf{0.902} & \textbf{0.978} & \textbf{0.833} & \textbf{270} \\
\bottomrule
\end{tabular}
}
\end{table}

The online feature-map AIN learns a binding-profile memory space in which simple probes recover a strong class structure after sequential exposure. This is evidence about the learned immune memory itself. Full refit k-means and replay MLP remain relevant upper baselines because they reuse old data to reconstruct or rehearse the decision space. The probe result answers whether the immune layers themselves produce a useful online visual memory space. The answer is positive in this diagnostic, especially because both a linear probe and a non-parametric 1NN probe remain strong after all ten digit classes have arrived.
For prototype-based rows, the final column reports the number of prototype or centroid memory slots. For feature-map Deep AIN rows, it reports the dimensionality of the final binding-profile embedding $z(X)$ used by the external probe, not the total number of B-cells across layers.

Table~\ref{tab:class_incremental_timing} reports training and inference timing for the same sklearn digits protocol. These values are intended as implementation-level diagnostics rather than hardware-independent complexity claims. For ordinary methods, training time includes initial fitting and all online updates. For feature-map Deep AIN rows, memory-training time reports immune-layer initialization and updates, while probe-fitting time reports the external validation probe fitted after each event.

\begin{table}[htbp]
\centering
\caption{Training and inference time for the sklearn digits Table~\ref{tab:class_incremental_final} protocol. Values are seconds averaged over three seeds. Inference time is the total time across all ten evaluations.}
\label{tab:class_incremental_timing}
\resizebox{\textwidth}{!}{%
\begin{tabular}{lrrrrrr}
\toprule
Method & Memory train & Probe fit & Total train & Total inference & Train/event & Inference/eval \\
\midrule
Online prototype memory & 0.008 & 0.000 & 0.008 & 0.129 & 0.001 & 0.0129 \\
Static k-means & 0.081 & 0.000 & 0.081 & 0.088 & 0.008 & 0.0088 \\
Mini-batch k-means & 0.165 & 0.000 & 0.165 & 0.083 & 0.017 & 0.0083 \\
No-replay MLP & 0.554 & 0.000 & 0.554 & 0.007 & 0.055 & 0.0007 \\
Replay MLP & 0.954 & 0.000 & 0.954 & 0.005 & 0.095 & 0.0005 \\
Full refit k-means & 1.525 & 0.000 & 1.525 & 0.126 & 0.152 & 0.0126 \\
Feature-map Deep AIN + 1NN probe & \textbf{4.385} & \textbf{0.008} & \textbf{4.393} & \textbf{0.022} & \textbf{0.439} & \textbf{0.0022} \\
Feature-map Deep AIN + logistic probe & \textbf{4.385} & \textbf{1.111} & \textbf{5.496} & \textbf{0.008} & \textbf{0.550} & \textbf{0.0008} \\
\bottomrule
\end{tabular}
}
\end{table}

Table~\ref{tab:mnist_incremental_final} repeats the same class-incremental protocol on a different dataset: MNIST. This is a more difficult visual setting because the images are larger and more variable. The feature-map Deep AIN again refers to the unsupervised immune layers; logistic regression and 1NN are external probes fitted on the learned binding-profile memory.

\begin{table}[htbp]
\centering
\caption{Final ten-class performance in the MNIST class-incremental visual-memory stream. Values are averaged over three seeds. Probe rows externally validate the learned Deep AIN binding-profile memory.}
\label{tab:mnist_incremental_final}
\resizebox{\textwidth}{!}{%
\begin{tabular}{llcccc}
\toprule
Method & Evaluation mode & Balanced accuracy & Initial retention & Current-class accuracy & Memory size / embedding dimension \\
\midrule
Static k-means & no update & 0.100 & 1.000 & 0.000 & 80 \\
Mini-batch k-means & online centroid update & 0.329 & 0.867 & 0.067 & 80 \\
Full refit k-means & refit on seen data & 0.699 & 0.933 & 0.756 & 120 \\
Online prototype memory & supervised online prototypes & 0.779 & 0.911 & 0.733 & 120 \\
No-replay MLP & current batch only & 0.444 & 0.100 & 0.967 & -- \\
Replay MLP & replay classifier & 0.851 & 0.900 & 0.856 & -- \\
Feature-map Deep AIN + logistic probe & external probe on $z(X)$ & \textbf{0.851} & \textbf{0.900} & \textbf{0.911} & \textbf{15870} \\
Feature-map Deep AIN + 1NN probe & external probe on $z(X)$ & \textbf{0.860} & \textbf{0.933} & \textbf{0.878} & \textbf{15870} \\
\bottomrule
\end{tabular}
}
\end{table}

The MNIST result sharpens the interpretation. The no-replay MLP achieves high current-class accuracy but loses the initial class almost completely, which is the expected forgetting pattern when no old data is rehearsed. Replay MLP avoids this collapse. The feature-map Deep AIN binding-profile memory is competitive with replay MLP under external probing, with the 1NN probe slightly higher in balanced accuracy and initial retention in this diagnostic. This supports the claim that the immune layers are learning a stable and coherent visual memory space.
As in Table~\ref{tab:class_incremental_final}, the final column reports memory slots for prototype rows and final binding-profile dimensionality for feature-map Deep AIN rows.

Table~\ref{tab:fashion_incremental_final} repeats the protocol on Fashion-MNIST using the same feature-map architecture and the same adaptive layer-wise affinity scale rule. This is an important transfer test because the images are still grayscale $28\times28$, but the visual categories are clothing classes rather than digit strokes.

\begin{table}[htbp]
\centering
\caption{Final ten-class performance in the Fashion-MNIST class-incremental visual-memory stream. Values are averaged over three seeds. Feature-map rows use the same adaptive layer-wise affinity scale rule.}
\label{tab:fashion_incremental_final}
\resizebox{\textwidth}{!}{%
\begin{tabular}{llcccc}
\toprule
Method & Evaluation mode & Balanced accuracy & Initial retention & Current-class accuracy & Memory size / embedding dimension \\
\midrule
Static k-means & no update & 0.100 & 1.000 & 0.000 & 80 \\
Mini-batch k-means & online centroid update & 0.469 & 0.700 & 0.767 & 80 \\
Full refit k-means & refit on seen data & 0.649 & 0.733 & 0.900 & 120 \\
Online prototype memory & supervised online prototypes & 0.651 & 0.833 & 0.844 & 120 \\
No-replay MLP & current batch only & 0.298 & 0.000 & 1.000 & -- \\
Replay MLP & replay classifier & 0.780 & 0.756 & 0.933 & -- \\
Feature-map Deep AIN + logistic probe & external probe on $z(X)$ & \textbf{0.814} & \textbf{0.811} & \textbf{0.922} & \textbf{15870} \\
Feature-map Deep AIN + 1NN probe & external probe on $z(X)$ & \textbf{0.801} & \textbf{0.811} & \textbf{0.944} & \textbf{15870} \\
\bottomrule
\end{tabular}
}
\end{table}

The Fashion-MNIST result is important because it was not obtained by selecting a new dataset-specific RBF sigma. The same adaptive layer-wise affinity rule produces a binding-profile space that is competitive with replay MLP and stronger than the rest of the baselines in this stream. Therefore, the feature-map Deep AIN does not just demonstrate single-dataset superiority; instead, it demonstrates general competitiveness under a fixed operating protocol.

Table~\ref{tab:cross_dataset_replay_ain} makes a direct comparison against replay MLP across the current visual datasets using the fixed adaptive scale rule. The result supports an important claim: a replay-free, gradient-free feature-map Deep AIN learns binding-profile memory spaces that are consistently competitive with a replay-based neural classifier under the same class-incremental protocol. Replay MLP retains old information by storing and rehearsing previous samples while the Deep AIN instead retains information through immune repertoires and the response-profile representation induced by those repertoires. KMNIST also reveals a useful diagnostic distinction: the 1NN probe is much stronger than the logistic probe, suggesting that the learned memory space can be locally class-coherent even when it is not linearly organized.

\begin{table}[htbp]
\centering
\caption{Cross-dataset comparison of replay MLP and feature-map Deep AIN binding-profile probes. Values are final ten-class performance averaged over three seeds. Deep AIN rows use the same adaptive layer-wise affinity scale rule on every dataset.}
\label{tab:cross_dataset_replay_ain}
\resizebox{\textwidth}{!}{%
\begin{tabular}{llccc}
\toprule
Dataset & Method & Balanced accuracy & Initial retention & Current-class accuracy \\
\midrule
Sklearn digits & Replay MLP & 0.946 & 0.989 & 0.944 \\
Sklearn digits & Feature-map Deep AIN + logistic probe & 0.943 & 0.978 & 0.933 \\
Sklearn digits & Feature-map Deep AIN + 1NN probe & 0.952 & 1.000 & 0.944 \\
MNIST & Replay MLP & 0.858 & 0.911 & 0.944 \\
MNIST & Feature-map Deep AIN + logistic probe & 0.862 & 0.900 & 0.856 \\
MNIST & Feature-map Deep AIN + 1NN probe & 0.857 & 0.922 & 0.889 \\
Fashion-MNIST & Replay MLP & 0.780 & 0.756 & 0.933 \\
Fashion-MNIST & Feature-map Deep AIN + logistic probe & 0.814 & 0.811 & 0.922 \\
Fashion-MNIST & Feature-map Deep AIN + 1NN probe & 0.801 & 0.811 & 0.944 \\
KMNIST & Replay MLP & 0.757 & 0.889 & 0.811 \\
KMNIST & Feature-map Deep AIN + logistic probe & 0.711 & 0.811 & 0.644 \\
KMNIST & Feature-map Deep AIN + 1NN probe & 0.853 & 0.933 & 0.844 \\
\bottomrule
\end{tabular}
}
\end{table}

The KMNIST discrepancy between logistic probing and 1NN probing motivates a more direct probe-geometry diagnostic. Table~\ref{tab:kmnist_probe_geometry} compares probes that test different forms of structure in the same learned binding-profile space. Nearest-centroid probing tests whether each class is represented by a compact global mean; logistic probing tests linear separability; PCA+RBF-SVM tests a simple nonlinear boundary after dimensionality reduction; and 1NN tests local neighbourhood coherence.

\begin{table}[htbp]
\centering
\caption{KMNIST probe-geometry diagnostic on the same learned feature-map Deep AIN binding-profile space. Values are final ten-class performance averaged over three seeds.}
\label{tab:kmnist_probe_geometry}
\begin{tabular}{lccc}
\toprule
Probe & Balanced accuracy & Initial retention & Current-class accuracy \\
\midrule
Nearest centroid & 0.612 & 0.811 & 0.589 \\
Logistic regression & 0.711 & 0.811 & 0.644 \\
PCA + RBF-SVM & 0.818 & 0.911 & 0.778 \\
1-nearest-neighbour & 0.853 & 0.933 & 0.844 \\
\bottomrule
\end{tabular}
\end{table}

This pattern supports the interpretation that the Deep AIN representation is meaningful but not necessarily globally linear. If the binding-profile space were simply failing, all probes would be weak. Instead, nonlinear and local probes substantially exceed linear and centroid probes. The evidence therefore suggests that feature-map Deep AINs can form locally coherent visual memory spaces whose quality may be underestimated by linear probes alone.

Table~\ref{tab:class_incremental_trace} shows the trajectory of the learned binding-profile space as classes are added.

\begin{table}[htbp]
\centering
\caption{External probe trajectory on online feature-map Deep AIN binding profiles in the sklearn digits class-incremental stream. Values are averaged over three seeds.}
\label{tab:class_incremental_trace}
\resizebox{\textwidth}{!}{%
\begin{tabular}{clcccc}
\toprule
Step & Encountered classes & LogReg balanced accuracy & 1NN balanced accuracy & LogReg initial retention & 1NN initial retention \\
\midrule
0 & $\{0\}$ & 1.000 & 1.000 & 1.000 & 1.000 \\
1 & $\{0,1\}$ & 1.000 & 1.000 & 1.000 & 1.000 \\
2 & $\{0,1,2\}$ & 0.981 & 0.985 & 1.000 & 1.000 \\
3 & $\{0,1,2,3\}$ & 0.953 & 0.944 & 0.989 & 1.000 \\
4 & $\{0,1,2,3,4\}$ & 0.971 & 0.962 & 0.989 & 0.978 \\
5 & $\{0,\ldots,5\}$ & 0.957 & 0.963 & 0.989 & 0.989 \\
6 & $\{0,\ldots,6\}$ & 0.951 & 0.944 & 0.989 & 0.989 \\
7 & $\{0,\ldots,7\}$ & 0.956 & 0.938 & 0.989 & 0.989 \\
8 & $\{0,\ldots,8\}$ & 0.943 & 0.905 & 0.989 & 0.989 \\
9 & $\{0,\ldots,9\}$ & 0.939 & 0.902 & 0.978 & 0.978 \\
\bottomrule
\end{tabular}
}
\end{table}

The decline in balanced accuracy as more classes arrive is expected: the representation task becomes harder with each new class. The key observation is that both probes remain strong and initial-class retention remains high, which indicates that the online immune layers have not collapsed the earlier class structure. The most difficult transition in this run occurs around digit 8, suggesting that class-specific visual overlap should be diagnosed per class rather than only by final average accuracy.

\subsection{Depth and Layer-Wise Scale Calibration}

Depth is useful only if later layers receive response spaces at a workable affinity scale. The feature-map AIN therefore requires a layer-wise scale choice for immune updates in binding-profile spaces. Table~\ref{tab:depth_sigma} summarizes the final ten-class sklearn digits result for one-, two-, and three-layer feature-map AINs. The depth-one model uses the first response maps directly as the validation embedding. Depth two and depth three learn over response-map spaces. The manually calibrated rows use fixed higher-layer training scales; the adaptive rows estimate the nearest-neighbour distance scale in the binding-profile space and multiply it by a scale factor.

\begin{table}[htbp]
\centering
\caption{Depth and layer-wise scale calibration in the sklearn digits class-incremental stream. Values are final-step means over three seeds. Probe rows externally validate the learned binding-profile memory.}
\label{tab:depth_sigma}
\resizebox{\textwidth}{!}{%
\begin{tabular}{llcccccc}
\toprule
Depth & Scale setting & Probe & Balanced accuracy & Initial retention & Current-class accuracy & Dimension & Interpretation \\
\midrule
1 & direct layer-1 maps & LogReg & 0.977 & 1.000 & 0.989 & 480 & strong shallow visual response space \\
1 & direct layer-1 maps & 1NN & 0.982 & 1.000 & 0.978 & 480 & strong shallow visual response space \\
2 & fixed $\sigma_2=1.0$ & LogReg & 0.953 & 0.978 & 0.967 & 270 & useful but under-scaled depth \\
2 & fixed $\sigma_2=2.0$ & LogReg & 0.978 & 0.978 & 1.000 & 270 & best calibrated two-layer result \\
2 & adaptive NN scale $\times 6$ & LogReg & 0.968 & 0.989 & 0.967 & 270 & principled scale rule, near manual best \\
3 & fixed $\sigma_2=2.0,\sigma_3=2.0$ & LogReg & 0.924 & 0.989 & 0.933 & 120 & third layer improves with scale but compresses strongly \\
3 & adaptive NN scales $\times 6,\times 6$ & LogReg & 0.927 & 0.989 & 0.944 & 120 & best current three-layer result \\
\bottomrule
\end{tabular}
}
\end{table}

The result is useful in two ways. First, it shows that poor higher-layer performance can be caused by scale mismatch rather than by depth itself. Increasing the second-layer training scale from 1.0 to 2.0 raises the two-layer logistic-probe result from 0.953 to 0.978 balanced accuracy. Second, it prevents an overclaim: deeper is not automatically better. A third layer retains the initial class well but compresses the representation from 270 to 120 dimensions and remains weaker than the calibrated two-layer model. This indicates that additional depth requires matched capacity and receptive-field choices rather than depth alone.

\section{Discussion}

The results support the central claim of the paper: visual AINs require structured affinity. The weakest results occur when images are treated as flattened vectors and compared by ordinary RBF affinity. This is not surprising, because a flattened representation makes the affinity score sensitive to pixel coordinate changes rather than to visual similarity. The improvement obtained by shifted-template matching and convolution-style ZNCC therefore has a clear interpretation. It shows that the first obstacle for visual AINs is not necessarily the immune learning mechanism itself, but the mismatch between image structure and the affinity operator. Once the B-cell is allowed to behave as a structured visual template, immune memory becomes a much more plausible mechanism for visual data.

The strongest evidence comes from the feature-map Deep AIN. Scalar binding profiles collapse each B-cell response into one number immediately. That can be appropriate for low-dimensional tabular antigens, but it removes too much information from visual antigens. The feature-map model delays this collapse: a layer-1 B-cell produces a spatial response map, and the next layer learns over the stack of response maps. This is important because it gives a concrete computational meaning to depth in a visual AIN. A deeper immune layer is not simply another classifier placed after the first layer; it is a repertoire that learns memories of lower-layer response patterns. In immune terms, the higher layer is exposed to the system's own response to the antigen, not only to the raw antigen itself.

The class-incremental results show that this response-pattern memory can remain useful under sequential exposure. The feature-map Deep AIN is not trained with labels and does not replay old samples. Nevertheless, its binding-profile space supports strong external probes after classes have arrived sequentially. This matters because the probes are not the proposed model; they are measurements of the representation produced by the AIN. The positive result therefore indicates that the immune repertoire has organized the visual stream into a memory space with recoverable class structure. This is the main empirical claim of the paper.

As new classes arrive, the lower immune layers adapt and the binding-profile coordinate system changes. The changing coordinate system is part of the online memory mechanism. The AIN is not replay-trained on old classes when new classes arrive. Instead, after immune adaptation, old and new samples can be passed through the current repertoire to obtain their current binding profiles. If earlier classes remain separable in this current representation space, then the system has \textbf{reorganized its representation} while retaining the ability to represent previous classes. The external readout is therefore used to test whether the current immune memory (representation) space still contains recoverable information about \text{all} encountered classes.

The comparison with replay MLP and full refit k-means should be interpreted carefully. Those methods are valid upper baselines for predictive performance because they reuse old information more directly: replay MLP rehearses stored examples, and full refit k-means reconstructs the clustering space from accumulated data. The feature-map Deep AIN is different in kind. Its immune layers adapt online, without replay, and form a binding-profile representation without backpropagation. Matching or exceeding replay MLP on some datasets is therefore significant, but the more important result is consistency: across sklearn digits, MNIST, Fashion-MNIST, and KMNIST, the learned immune representation remains competitive enough to be useful. This supports the claim that structured-affinity AINs are viable visual online memory learners, not merely small-dataset curiosities.

The probe results also reveal the geometry of the learned memory space. On KMNIST, 1-nearest-neighbour and PCA+RBF-SVM outperform logistic regression and nearest-centroid probes. This suggests that the binding-profile space can be locally coherent without being globally linearly separable. That observation is important for two reasons. First, it prevents an overly narrow evaluation of AIN representations using only linear probes. Second, it suggests that future immune readout mechanisms should not assume that useful memory must appear as a globally linear class structure. Local neighbourhoods, response-pattern similarity, and repertoire-level interactions may be more faithful to the immune interpretation.

The depth and scale-calibration results further qualify the role of depth. Depth is useful only when the higher layer operates at an appropriate affinity scale. A poorly scaled response-profile space can make a deeper AIN appear weak even when the underlying representation is meaningful. The performance improvement from fixed under-scaled settings to calibrated second-layer settings shows that higher-layer B-cells require scale selection appropriate to the response-profile space, not to the original image space. At the same time, the three-layer result shows that adding depth alone is not sufficient. Additional layers compress and reorganize the representation, so their capacity and receptive fields must be matched to the structure of the previous layer's response maps.

Overall, the paper establishes a specific claim. It does not show that AINs outperform modern deep visual classifiers, nor does it claim that ZNCC is the best possible immune affinity. Rather, it shows that replay-free visual online learning in AINs becomes credible when three conditions hold: B-cells use structured visual affinity, lower-layer responses preserve spatial response maps, and higher layers are calibrated to the binding-profile spaces they receive. Under those conditions, a Deep AIN can form a replay-free, gradient-free visual memory representation whose class structure is recoverable after sequential exposure.

\section{Limitations}

The results support structured-affinity Deep AINs as replay-free visual online memory learners, but they also identify several limitations of the current formulation.

\textbf{Local-filter ambiguity.}
The feature-map model relies on local B-cell filters. Local responses are useful because they preserve spatial structure and make the AIN less dependent on flattened pixel distance. However, small local filters can also respond similarly to visually related classes. This is especially likely in digit and character data, where different classes may share strokes, corners, loops, and short edge fragments. In these cases, a lower layer may produce response maps that are locally meaningful but not sufficiently distinctive for global class discrimination. The KMNIST probe diagnostic is consistent with this issue: the learned space remains locally coherent, but linear and centroid-based probes are weaker than local or nonlinear probes.

\textbf{Repertoire growth and redundancy.}
Memory formation in an AIN can increase the number of B-cells or the dimensionality of the binding-profile representation. This is conceptually attractive because adding B-cells adds memory axes, but it also creates a practical tradeoff. Larger repertoires increase training cost, inference cost, and the dimensionality of downstream binding profiles. They may also accumulate redundant or low-utility B-cells if suppression does not remove them effectively. The present paper reports runtime diagnostics, but it does not yet provide a full population-control mechanism for preserving compact repertoires while maintaining visual memory quality.

\textbf{Higher-layer scale and capacity sensitivity.}
The depth experiments show that higher-layer performance depends strongly on the affinity scale and capacity of the response-profile space. A second layer improves substantially when its scale is calibrated, whereas a third layer remains weaker under the current configuration because it compresses the representation too aggressively. This does not undermine the definition of depth, but it shows that additional immune layers require matched receptive fields, repertoire sizes, and scale-selection rules. Depth is therefore a mechanism that must be configured with respect to the geometry of the previous layer's response maps.

\textbf{Scope of the current visual evidence.}
The strongest evidence in this paper is obtained on grayscale visual benchmarks: sklearn digits, MNIST, Fashion-MNIST, and KMNIST. These datasets are appropriate for establishing the core claim that structured affinity makes online visual AIN learning viable, but they do not exhaust the visual setting. Natural RGB images introduce colour channels, stronger background variation, texture diversity, and more complex object structure. The present results therefore establish a proof of concept for structured visual immune memory rather than a general solution to visual continual learning.

\section{Conclusion}

This paper provides evidence that AINs can learn visual data online when affinity respects image structure. Flattened vector affinity is a weak inductive bias for visual antigens, but structured immune affinity allows B-cells to act as local visual templates and response-map generators. In this setting, a Deep AIN is not a supervised classifier; it is a replay-free, gradient-free immune representation-memory learner whose binding-profile space can be evaluated by external probes.

The strongest evidence is that feature-map Deep AIN binding profiles form class-discriminative memory spaces under sequential visual exposure. On sklearn digits, a calibrated two-layer feature-map Deep AIN reaches 0.978 final balanced accuracy with a logistic external probe. Under the same adaptive scale rule, the model remains competitive across MNIST, Fashion-MNIST, and KMNIST, with the KMNIST diagnostic indicating that the learned memory space can be locally coherent even when it is not globally linearly separable. These results support the paper's central claim: structured affinity makes visual online memory learning a viable and scientifically meaningful direction for artificial immune networks.

\bibliographystyle{plainnat}
\bibliography{references}

@article{jerne1974network,
  author = {Jerne, Niels K.},
  title = {Towards a Network Theory of the Immune System},
  journal = {Annales d'Immunologie},
  volume = {125C},
  number = {1--2},
  pages = {373--389},
  year = {1974}
}

@article{farmer1986immune,
  author = {Farmer, J. Doyne and Packard, Norman H. and Perelson, Alan S.},
  title = {The Immune System, Adaptation, and Machine Learning},
  journal = {Physica D: Nonlinear Phenomena},
  volume = {22},
  number = {1--3},
  pages = {187--204},
  year = {1986}
}

@book{decastro2002ais,
  author = {de Castro, Leandro N. and Timmis, Jonathan},
  title = {Artificial Immune Systems: A New Computational Intelligence Approach},
  publisher = {Springer},
  year = {2002}
}

@article{broomhead1988rbf,
  author = {Broomhead, David S. and Lowe, David},
  title = {Multivariable Functional Interpolation and Adaptive Networks},
  journal = {Complex Systems},
  volume = {2},
  pages = {321--355},
  year = {1988}
}

@book{bishop1995nn,
  author = {Bishop, Christopher M.},
  title = {Neural Networks for Pattern Recognition},
  publisher = {Oxford University Press},
  year = {1995}
}

@book{pekalska2005dissimilarity,
  author = {Pekalska, Elzbieta and Duin, Robert P. W.},
  title = {The Dissimilarity Representation for Pattern Recognition: Foundations and Applications},
  publisher = {World Scientific},
  year = {2005}
}

@inproceedings{williams2001nystrom,
  author = {Williams, Christopher K. I. and Seeger, Matthias},
  title = {Using the Nystr{\"o}m Method to Speed Up Kernel Machines},
  booktitle = {Advances in Neural Information Processing Systems},
  year = {2001}
}

@book{dasgupta1999ais,
  editor = {Dasgupta, Dipankar},
  title = {Artificial Immune Systems and Their Applications},
  publisher = {Springer},
  year = {1999}
}

@inproceedings{forrest1994negative,
  author = {Forrest, Stephanie and Perelson, Alan S. and Allen, Lawrence and Cherukuri, Rajesh},
  title = {Self-Nonself Discrimination in a Computer},
  booktitle = {Proceedings of the 1994 IEEE Symposium on Research in Security and Privacy},
  pages = {202--212},
  year = {1994}
}

@article{decastro2002clonal,
  author = {de Castro, Leandro N. and Von Zuben, Fernando J.},
  title = {Learning and Optimization Using the Clonal Selection Principle},
  journal = {IEEE Transactions on Evolutionary Computation},
  volume = {6},
  number = {3},
  pages = {239--251},
  year = {2002}
}

@inproceedings{watkins2002airs,
  author = {Watkins, Andrew and Timmis, Jonathan},
  title = {Artificial Immune Recognition System ({AIRS}): An Immune-Inspired Supervised Learning Algorithm},
  booktitle = {Proceedings of the Genetic and Evolutionary Computation Conference},
  pages = {1390--1397},
  year = {2002}
}

@inproceedings{decastro2000ainet,
  author = {de Castro, Leandro N. and Von Zuben, Fernando J.},
  title = {An Evolutionary Immune Network for Data Clustering},
  booktitle = {Proceedings of the IEEE Brazilian Symposium on Artificial Neural Networks},
  pages = {84--89},
  year = {2000}
}

@article{timmis2000resource,
  author = {Timmis, Jonathan and Neal, Mark and Hunt, John},
  title = {An Artificial Immune System for Data Analysis},
  journal = {Biosystems},
  volume = {55},
  number = {1--3},
  pages = {143--150},
  year = {2000}
}

@article{kohonen1982som,
  author = {Kohonen, Teuvo},
  title = {Self-Organized Formation of Topologically Correct Feature Maps},
  journal = {Biological Cybernetics},
  volume = {43},
  number = {1},
  pages = {59--69},
  year = {1982}
}

@article{kohonen1990lvq,
  author = {Kohonen, Teuvo},
  title = {Improved Versions of Learning Vector Quantization},
  journal = {Proceedings of the International Joint Conference on Neural Networks},
  pages = {545--550},
  year = {1990}
}

@inproceedings{rahimi2007random,
  author = {Rahimi, Ali and Recht, Benjamin},
  title = {Random Features for Large-Scale Kernel Machines},
  booktitle = {Advances in Neural Information Processing Systems},
  year = {2007}
}

@article{kirkpatrick2017ewc,
  author = {Kirkpatrick, James and Pascanu, Razvan and Rabinowitz, Neil and Veness, Joel and Desjardins, Guillaume and Rusu, Andrei A. and Milan, Kieran and Quan, John and Ramalho, Tiago and Grabska-Barwinska, Agnieszka and Hassabis, Demis and Clopath, Claudia and Kumaran, Dharshan and Hadsell, Raia},
  title = {Overcoming Catastrophic Forgetting in Neural Networks},
  journal = {Proceedings of the National Academy of Sciences},
  volume = {114},
  number = {13},
  pages = {3521--3526},
  year = {2017}
}

@inproceedings{rebuffi2017icarl,
  author = {Rebuffi, Sylvestre-Alvise and Kolesnikov, Alexander and Sperl, Georg and Lampert, Christoph H.},
  title = {{iCaRL}: Incremental Classifier and Representation Learning},
  booktitle = {Proceedings of the IEEE Conference on Computer Vision and Pattern Recognition},
  pages = {2001--2010},
  year = {2017}
}

@inproceedings{lopezpaz2017gem,
  author = {Lopez-Paz, David and Ranzato, Marc'Aurelio},
  title = {Gradient Episodic Memory for Continual Learning},
  booktitle = {Advances in Neural Information Processing Systems},
  year = {2017}
}

@article{fukushima1980neocognitron,
  author = {Fukushima, Kunihiko},
  title = {Neocognitron: A Self-Organizing Neural Network Model for a Mechanism of Pattern Recognition Unaffected by Shift in Position},
  journal = {Biological Cybernetics},
  volume = {36},
  number = {4},
  pages = {193--202},
  year = {1980}
}

@article{lecun1998gradient,
  author = {LeCun, Yann and Bottou, L{\'e}on and Bengio, Yoshua and Haffner, Patrick},
  title = {Gradient-Based Learning Applied to Document Recognition},
  journal = {Proceedings of the IEEE},
  volume = {86},
  number = {11},
  pages = {2278--2324},
  year = {1998}
}

@inproceedings{lewis1995fast,
  author = {Lewis, J. P.},
  title = {Fast Normalized Cross-Correlation},
  booktitle = {Vision Interface},
  pages = {120--123},
  year = {1995}
}

@article{lowe2004distinctive,
  author = {Lowe, David G.},
  title = {Distinctive Image Features from Scale-Invariant Keypoints},
  journal = {International Journal of Computer Vision},
  volume = {60},
  number = {2},
  pages = {91--110},
  year = {2004}
}

@inproceedings{dalal2005hog,
  author = {Dalal, Navneet and Triggs, Bill},
  title = {Histograms of Oriented Gradients for Human Detection},
  booktitle = {Proceedings of the IEEE Conference on Computer Vision and Pattern Recognition},
  volume = {1},
  pages = {886--893},
  year = {2005}
}

@article{parisi2019continual,
  author = {Parisi, German I. and Kemker, Ronald and Part, Jose L. and Kanan, Christopher and Wermter, Stefan},
  title = {Continual Lifelong Learning with Neural Networks: A Review},
  journal = {Neural Networks},
  volume = {113},
  pages = {54--71},
  year = {2019}
}

@article{delange2021continual,
  author = {De Lange, Matthias and Aljundi, Rahaf and Masana, Marc and Parisot, Sarah and Jia, Xu and Leonardis, Ales and Slabaugh, Gregory and Tuytelaars, Tinne},
  title = {A Continual Learning Survey: Defying Forgetting in Classification Tasks},
  journal = {IEEE Transactions on Pattern Analysis and Machine Intelligence},
  volume = {44},
  number = {7},
  pages = {3366--3385},
  year = {2021}
}

@article{vandeven2022continual,
  author = {van de Ven, Gido M. and Tuytelaars, Tinne and Tolias, Andreas S.},
  title = {Three Types of Incremental Learning},
  journal = {Nature Machine Intelligence},
  volume = {4},
  pages = {1185--1197},
  year = {2022}
}

\end{document}